\documentclass{article}
\usepackage{spconf,amsmath,graphicx}
\usepackage{booktabs}
\usepackage{pifont}
\usepackage{amssymb}
\usepackage{caption}
\usepackage{multirow}
\usepackage{enumitem}
\usepackage{pdfpages}
\usepackage{subcaption}

\setlist{nosep, leftmargin=14pt}

\title{Fairness Analysis of CLIP-Based Foundation Models for X-ray Image Classification}

\name{Xiangyu Sun$^1$, Xiaoguang Zou$^{2,3}$, Yuanquan Wu$^{3,4}$, Guotai Wang$^{1,5*}$, Shaoting Zhang$^{1,5}$%
\thanks{*Corresponding author (\texttt{guotai.wang@uestc.edu.cn})}}

\address{
$^1$ School of Mechanical and Electrical Engineering\\
University of Electronic Science and Technology of China, Chengdu, China\\
$^2$ Clinical Medical Research Center, The First People’s Hospital of Kashi (Kashgar) Prefecture, China\\
$^3$ Xinjiang Key Laboratory of Artificial Intelligence Assisted Imaging Diagnosis, Kashi (Kashgar), China\\
$^4$ Department of Hepatobiliary Surgery, The First People’s Hospital of Kashi (Kashgar) Prefecture, China\\
$^5$ Shanghai Artificial Intelligence Laboratory, Shanghai, China\\
}

\begin{document}
\maketitle

\begin{abstract}
X-ray imaging is pivotal in medical diagnostics, offering non-invasive insights into a range of health conditions. Recently, vision-language models, such as the Contrastive Language-Image Pretraining (CLIP) model, have demonstrated potential in improving diagnostic accuracy by leveraging large-scale image-text datasets. However, since CLIP was not initially designed for medical images, several CLIP-like models trained specifically on medical images have been developed. Despite their enhanced performance, issues of fairness—particularly regarding demographic attributes remain largely unaddressed. In this study, we perform a comprehensive fairness analysis of CLIP-like models applied to X-ray image classification. We assess their performance and fairness across diverse patient demographics and disease categories using zero-shot inference and various fine-tuning techniques, including Linear Probing, Multilayer Perceptron (MLP), Low-Rank Adaptation (LoRA), and full fine-tuning. Our results indicate that while fine-tuning improves model accuracy, fairness concerns persist, highlighting the need for further fairness interventions in these foundational models.
\end{abstract}

\begin{keywords}
CLIP, X-ray image, Fairness, Fine-tuning
\end{keywords}

\section{Introduction}
Medical imaging, particularly X-ray imaging, plays a crucial role in diagnosing a variety of diseases. 
The integration of Artificial Intelligence (AI) into medical imaging has shown potential in enhancing diagnostic accuracy and efficiency~\cite{ccalli2021deep}. Deep learning models ~\cite{esteva2017dermatologist,qin2018computer,10635182,9761426,10635205} trained through fully supervised learning have achieved significant success in various medical image classification tasks. However, these models are constrained by high annotation costs and limited generalizability to new image types or datasets, which restricts their applicability across diverse clinical settings. 

Recent advancements in vision-language models, such as CLIP~\cite{radford2021learning}, have enhanced the classification of previously unseen images by leveraging large-scale image-text pairs for training and zero-shot inference. This approach is particularly appealing for medical image classification, where obtaining sufficient annotated training data is challenging and time-consuming. However, CLIP was originally trained primarily on natural image-text pairs, limiting its performance on medical images like X-rays that exhibit different semantic and visual distributions~\cite{8c537381a4af42e99177c9f1a94dc4cd}. To address this limitation, domain-adapted CLIP variants—GLoRIA~\cite{huang2021gloria}, MedCLIP~\cite{8c537381a4af42e99177c9f1a94dc4cd}, and BioMedCLIP~\cite{zhang2023biomedclip}—have been developed by training on extensive medical image-text datasets, resulting in improved performance in X-ray image classification.

Despite these advancements, fairness—ensuring consistent model performance across diverse patient demographics—remains inadequately addressed. Fairness in medical AI is essential to prevent biased model outputs that may lead to unequal healthcare outcomes, particularly concerning demographic factors such as age and gender~\cite{10230368}. Previous studies have demonstrated that biases can emerge at various stages of the model development pipeline, including pretraining and fine-tuning~\cite{jin2024fairmedfm}. Therefore, a comprehensive evaluation of fairness in these adapted CLIP-based models is crucial to ensure both accurate and equitable healthcare delivery.

\begin{figure*}[htbp]
    \centering
    \includegraphics[width=0.8\textwidth]{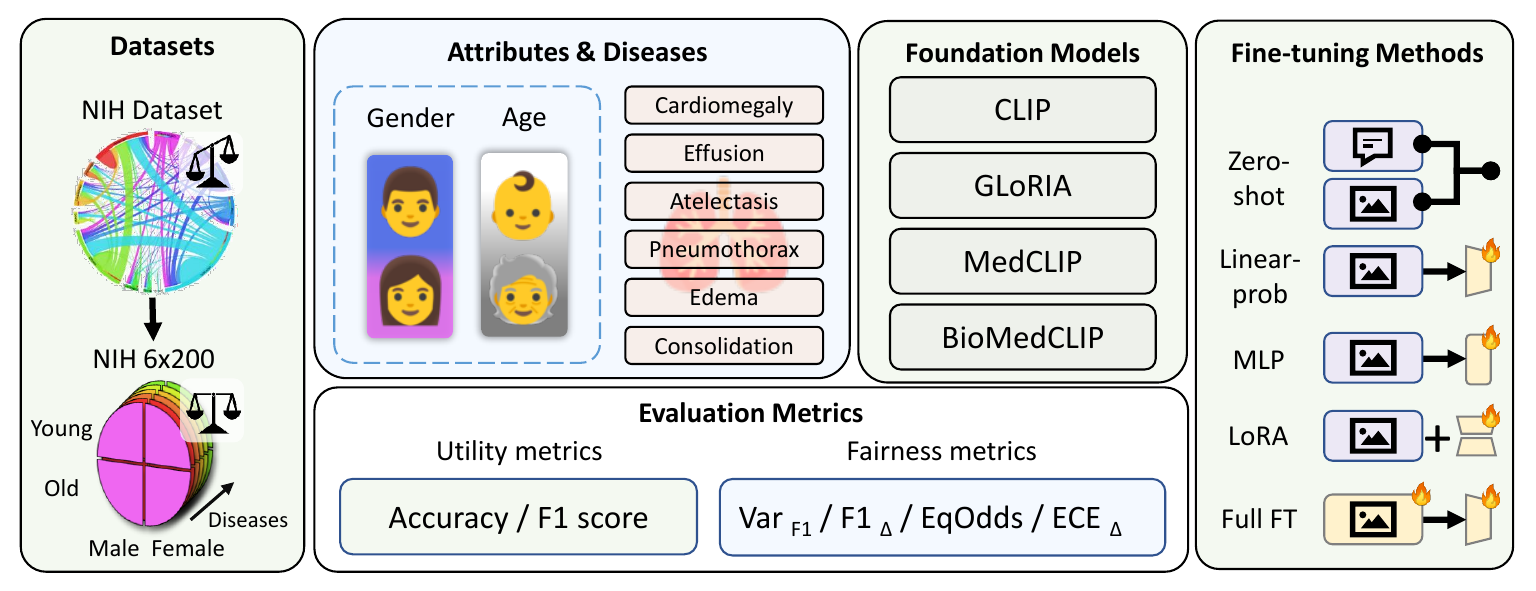} 
    \caption{Overview of the fairness analysis of different foundation models in X-ray image classification conducted in this study.} 
    \label{fig} 
\end{figure*}

In this work, we systematically evaluate the performance and fairness of medical CLIP models, focusing on their ability to generalize across different patient demographics and disease categories. An overview of this work is shown in Fig.~\ref{fig}, and the main contributions of this research include: 1) Creating a gender and age-balanced X-ray dataset with multiple diseases for fairness evaluation of foundation models; 2) Conducting a comprehensive evaluation of the fairness of four CLIP-like models on two sensitive attributes, including gender and age; 3) Investigating the effect of different fine-tuning methods on the fairness of these models. Our findings show that while fine-tuning improves classification accuracy, significant fairness gaps related to age and gender remain, underscoring the need for advanced bias mitigation strategies.

\section{Method}

\subsection{X-ray Dataset with Balanced Demographic Groups}
To evaluate the fairness and efficacy of CLIP-based models in X-ray image classification, we employed the NIH Chest X-ray dataset, which comprises 108,948 images from 32,717 patients, spanning fourteen distinct disease categories~\cite{wang2017chestx}. This dataset includes annotations for key demographic attributes such as age and gender. However, as the original dataset has large imbalanced distributions of age, gender and disease type, directly using it for fine-tuning may introduce additional unfairness. Therefore, we curated a subset with balanced demographic representation to facilitate fairness studies.

We considered gender and age as the sensitive attributes in our work. For each disease, we considered four patients groups: Young (age $<$ 60) Male, Old (age $>=$ 60) Male, Young Female, and Old Female. To avoid imbalance from the dataset, for each disease, we selected $N$ images from each patient group, leading to $4N$ images for each disease. In this study, we set $N = 50$ and found that only six diseases satisfy the data selection criteria: the image number in any of the four groups should be larger than $N$. Setting $N > 50$  will lead to reduced disease types, while $N < 50$ will make the sample number too small in each patient group. As a result, we selected six diseases each with $4\times50 = 200$ images, and name it as NIH $6\times200$ dataset. The disease types are: Cardiomegaly, Effusion, Atelectasis, Pneumothorax, Edema, and Consolidation. Note that an exception is the Edema class in the Old age group and it contains 47 female and 53 male cases respectively due to limitations in the available NIH data. This balanced dataset facilitates a rigorous evaluation of model fairness across different demographic groups.

\subsection{Models Under Investigation}
We investigate four CLIP-based models for X-ray image classification:
 1) CLIP~\cite{radford2021learning} that was trained on 400 million natural image-text pairs from the internet. It offers various image encoder architectures, and we utilize the ViT-B/16 model in our experiments. For text encoding, CLIP employs a standard BERT-based Transformer encoder.
 2) GLoRIA~\cite{huang2021gloria} that was trained on the CheXpert~\cite{irvin2019chexpert} dataset using a ResNet-50 image encoder. For text encoding, it employs BioClinicalBERT~\cite{alsentzer-etal-2019-publicly} pretrained on the MIMIC-III~\cite{johnson2016mimic} dataset.  
 3) MedCLIP~\cite{8c537381a4af42e99177c9f1a94dc4cd} that was trained on a combination of MIMIC-CXR~\cite{johnson2019mimic} and CheXpert datasets. It has two variants that use SwinTransformer and ResNet-50, respectively, denoted as MedCLIP$^{ViT}$ and MedCLIP$^{RN}$. MedCLIP also utilizes BioClinicalBERT for text encoding.  
 4) BioMedCLIP~\cite{zhang2023biomedclip} that was trained on the 
 PMC-15M dataset (100 times larger than MIMIC-CXR). It employs a Vision Transformer (ViT) as the image encoder and utilizes PubMedBERT~\cite{gu2021domain} for text encoding with refinements to the tokenizer and context size. It is important to note that none of these models was trained on the NIH Chest X-ray dataset. Therefore, evaluating these models on our NIH 6$\times$200 dataset ensures an unbiased assessment.

\begin{figure*}[htbp]
\centering
\begin{subfigure}[b]{0.48\textwidth}
    \centering
    \includegraphics[width=\textwidth]{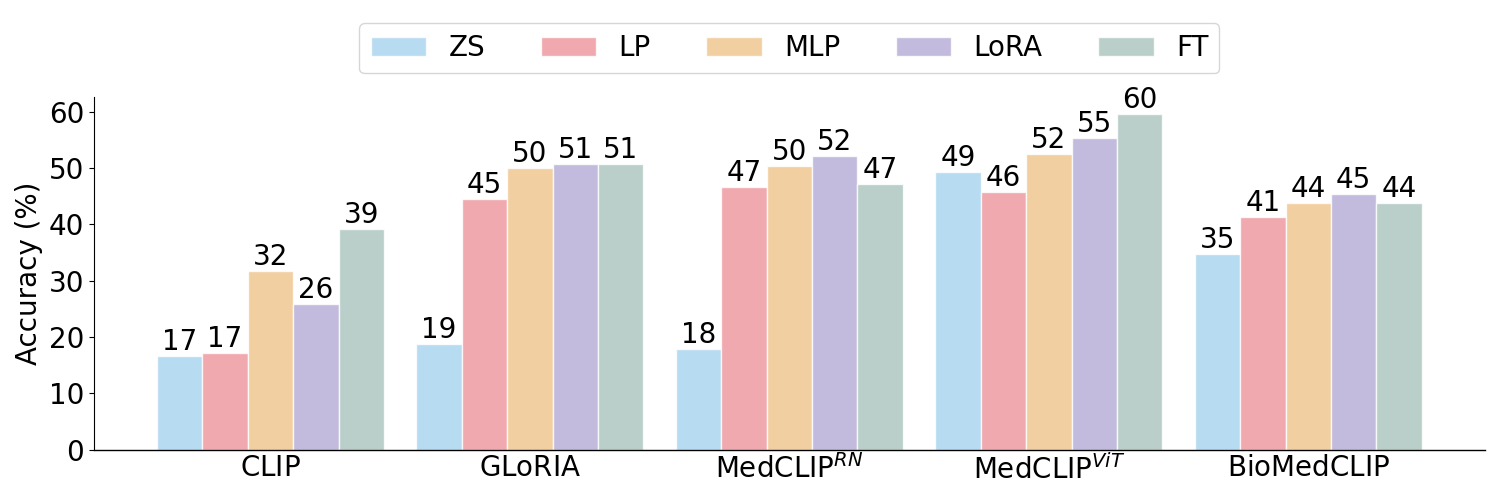} 
    \caption{Overall accuracy of each model}
    \label{figure1} 
\end{subfigure}
\hfill
\begin{subfigure}[b]{0.48\textwidth}
    \centering
    \includegraphics[width=\textwidth]{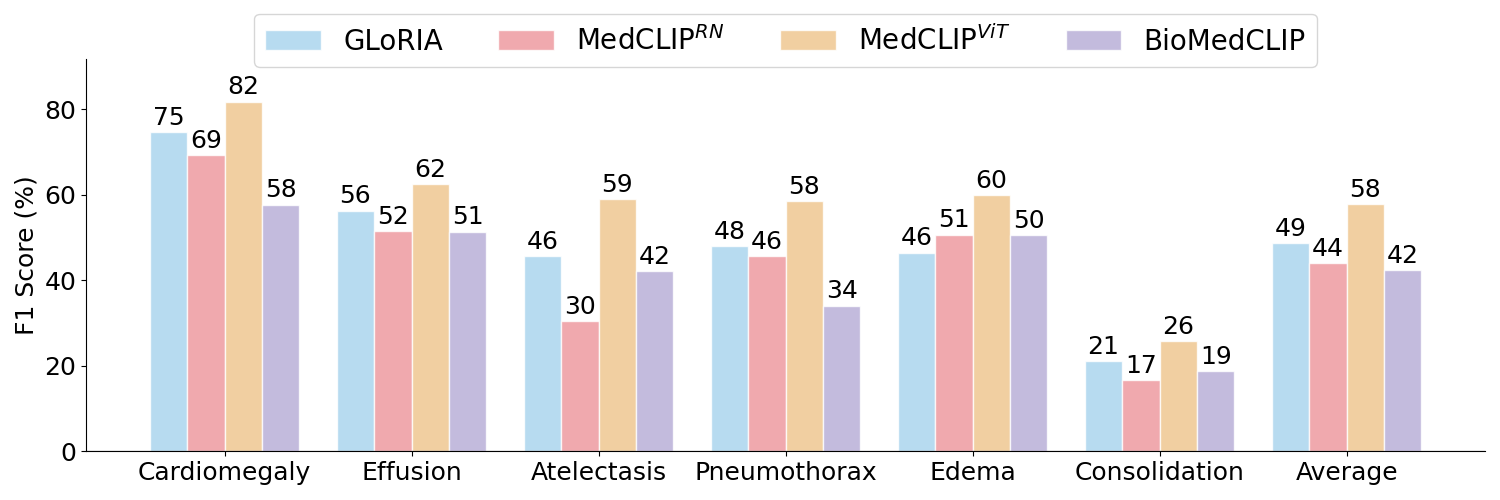}
    \caption{F1 Score for each disease category after full fine-tuning.}
    \label{figure2} 
\end{subfigure}
\hfill

\caption{Performance of various models under different fine-tuning configurations.}
\label{fig:performance_plots}
\end{figure*}

\subsection{Zero-shot Inference and Fine-Tuning Strategies}
We first evaluate the models' generalization capabilities using Zero-Shot (ZS) Inference, which involves directly applying the pretrained models with text prompts without fine-tuning. Specifically, we employ three distinct prompt formats as proposed in the CLIP, GLoRIA, and CXR-CLIP studies~\cite{radford2021learning, huang2021gloria, you2023cxr}. Since the latter two methods involve random prompt combinations, for each prompt format, we took an average of 10 runnings and selected the best-performing prompt format for each model.

To further adapt the CLIP-like models for X-ray image classification, we employ four fine-tuning strategies:
 1) Linear Probing (LP) that freezes the pretrained backbone and trains only a single-layer classification head, allowing minimal adaptation while preserving learned features.
 2) Multilayer Perceptron (MLP) that introduces an additional nonlinear layer between the frozen backbone and the output classification layer to learn complex representations.
 3) Low-Rank Adaptation (LoRA)~\cite{hu2022lora} that applies low-rank updates to all layers of the image encoder for parameter-efficient fine-tuning.
 4) Full Fine-Tuning (FT) that updates the entire image encoder to enable complete adaptation to the target task. Note that for CLIP~\cite{radford2021learning}, fine-tuning the entire image encoder was not allowed due to the hardware limitation, we used a ViT pretrained on ImageNet for fine-tuning instead.

\begin{figure*}[htbp]
    \centering
    \includegraphics[width=\textwidth]{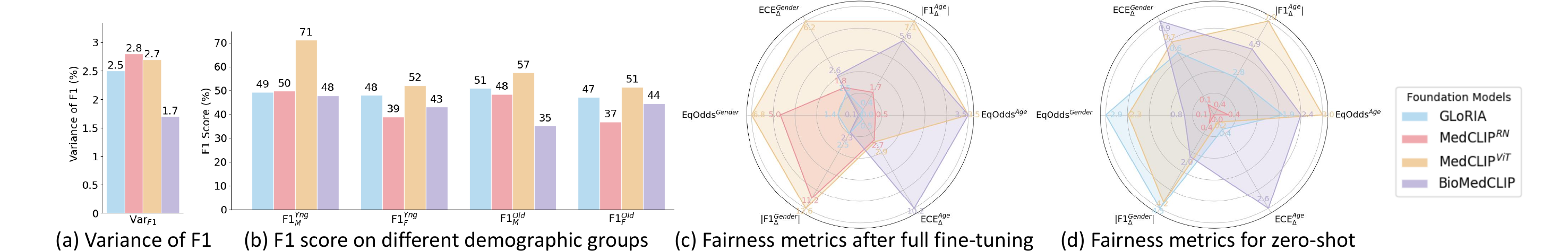} 
    \caption{Fairness analysis after full fine-tuning. (a) Fairness among diffident disease types in terms of variance of F1. (b) shows the F1-score of different demographic groups, and (c) and (d) show fairness metrics for age and gender.} 
    \label{fig:fairness} 
\end{figure*}

\subsection{Utility and Fairness Metrics}
To comprehensively assess models’ performance, we apply both utility and fairness metrics.

\textbf{Utility Metrics}: Three metrics are used to evaluate the models' performance: {1) Accuracy} that measures the overall proportion of correct predictions; {2) Class-wise F1 score} that is the harmonic mean of precision and recall for each class; and {3) Average F1-score} for each demographic group. 

\textbf{Fairness Metrics}:
Firstly, to evaluate the fairness across different disease types, we measured the variance of F1 score ($\text{Var}_{F1}$) across all the disease categories, see Equation~\ref{eq:var_f1}.

\begin{equation}
\text{Var}_{F1} = \frac{1}{C} \sum_{c=1}^{C} (F1_c - \overline{F1})^2
\label{eq:var_f1}
\end{equation}

where \(C\) is the number of disease categories, \(F1_c\) is the F1 score for the \(c\)-th category, and \(\overline{F1}\) is the mean F1 score for the \(C\) classes.

Secondly, for the fairness assessment on gender and age, we employ: 1) {Gap of F1 Score ($\text{F1}_{\Delta}$)} that represents the average difference in F1 scores between demographic groups across all classes; 2) {Equalized Odds (EqOdds)} that combines gaps of both true positives and false positives across demographic groups; and 3) {Gap of Expected Calibration Error ($\text{ECE}_{\Delta}$)} that quantifies the difference in model calibration across demographic groups~\cite{jin2024fairmedfm}. See Equations~\ref{eq:delta_f1}, \ref{eq:eqodds}, and \ref{eq:delta_ece}.

\begin{equation}
\text{F1}_{\Delta} = \frac{1}{C} \sum_{c=1}^{C} \left( \text{F1}_{A}^{c} - \text{F1}_{A'}^{c} \right)
\label{eq:delta_f1}
\end{equation}

where \(\text{F1}_{A}^{c}\) and \(\text{F1}_{A'}^{c}\) are the F1 scores for class \(c\) in demographic groups \(A\) and \(A'\) (e.g., male and female), respectively.

\begin{equation}
\text{EqOdds} = \frac{1}{2C} \sum_{c=1}^{C} \left( |\text{TP}_{A}^{c} - \text{TP}_{A'}^{c}| + |\text{FP}_{A}^{c} - \text{FP}_{A'}^{c}| \right)
\label{eq:eqodds}
\end{equation}

where \(\text{TP}_{A}^{c}\), \(\text{FP}_{A}^{c}\) and \(\text{TP}_{A'}^{c}\), \(\text{FP}_{A'}^{c}\) are the true positive, false positive rates for class \(c\) in demographic group \(A\) and \(A'\), respectively.

\begin{equation}
\text{ECE}_{\Delta} = \left| \frac{1}{N_A} \sum_{i=1}^{N_A} | p_i - o_i | - \frac{1}{N_{A'}} \sum_{j=1}^{N_{A'}} | p_j' - o_j' | \right|
\label{eq:delta_ece}
\end{equation}

where \(N_A\) and \(N_{A'}\) are the number of samples in the two demographic groups, respectively. 
\(p_i\) and \(o_i\) are the predicted probability and actual outcome for sample \(i\) in the first group, and \(p_j'\) and \(o_j'\) are those for sample \(j\) in the second group.

\section{Experiments and Results}

\subsection{Implementation Details}

We utilized the NIH 6$\times$200 dataset to ensure a balanced representation across six distinct disease categories and two demographic attributes: age and gender. Our NIH 6x200 dataset was partitioned into training, validation, and testing sets at a ratio of 7:1:2 to enable robust model assessment. 



All experiments were performed on a single NVIDIA GTX 1080Ti GPU. Each model was reproduced and fine-tuned within the software environments specified in their original publications. For model fine-tuning, we employed the AdamW optimizer with a cosine annealing learning rate scheduler, with a batch size of 64, and trained for 150 epochs. Other hyperparameters, such as the learning rate, were individually adjusted based on validation performance. We retained the checkpoint with the lowest validation loss for evaluation on the testing set.

\subsection{Utility  Analysis}

As presented in Fig.~\ref{fig:performance_plots}(a), the original CLIP model had the lowest classification performance in both zero-shot inference and fine-tuning settings, due to that the other models were specifically trained for medical images. 
In terms of zero-shot inference, MedCLIP$^{ViT}$ outperformed the other models with a classification accuracy of 49.4\%. 
However, the performance remains limited, highlighting the challenges of applying pretrained models directly to X-ray image classification tasks without adaptation. 

In addition, Fig.~\ref{fig:performance_plots}(a) shows that all the fine-tuning methods led to performance improvement. Full fine-tuning largely outperformed the other fine-tuning strategies for CLIP and MedCLIP$^{ViT}$, and LoRA obtained the best performance on MedCLIP$^{RN}$ and BioMedCLIP.
Among all these models and fine-tuning strategies, MedCLIP using a ViT architecture with full fine-tuning achieved the highest overall accuracy of 59.6\%,  highlighting the effectiveness fine-tuning in enhancing the model's performance.
Fig.~\ref{fig:performance_plots}(b) shows the class-wise F1 score of different models after full fine-tuning. It can be observed that MedCLIP$^{ViT}$ consistently outperformed the other models across all the evaluated disease categories. 


\subsection{Fairness Analysis}
\textbf{Fairness on Disease Types}. A large variation of F1-score among different disease categories can be found in Fig.~\ref{fig:performance_plots}(b). For example, the class-wise F1-score of GLoRIA ranged from 21.1\% to 74.7\%, while that for MedCLIP$^{ViT}$ ranged from 25.8\% to 81.8\%, demonstrating the obvious unfairness of these models on different disease types. 
Fig.~\ref{fig:fairness}(a) further compares the variance of class-wise F1-score of these methods, and it shows  that MedCLIP$^{RN}$ and  MedCLIP$^{ViT}$ have  very close Var$_{F1}$ values, and they are much  larger than those of GLoRIA and BioMedCLIP.
Generally,  BioMedCLIP has the highest disease-level fairness (Var$_{F1}$=1.7\%), in despite of its lower classification performance than MedCLIP$^{ViT}$.  

\textbf{Fairness on Age and Gender}. Fig. ~\ref{fig:fairness}(b) shows the average class-wise F1-score on different demographic groups. It can be observed that MedCLIP$^{VIT}$ has a large unfairness between young male and young female ({71.1\% vs 52.0\%} in terms of F1), and the gap between young male and old male is also large ({71.1\% vs 57.4\%}). In contrast,  GLoRIA exhibits a relatively fair performance on these demographic groups with F1 ranging from {47.1\% to 50.9\%}. 
Fig.~\ref{fig:fairness}(c) presents more fairness metrics of these models after full fine-tuning including $\text{F1}_{\Delta}$, EqOdds and $\text{ECE}_{\Delta}$ for age and gender. 
It is evident that all models, except GLoRIA, exhibit relatively high fairness metric values, indicating significant disparities across different demographic groups.
Specifically, MedCLIP$^{\text{ViT}}$ has the highest $\text{F1}_{\Delta}^{\text{Gender}}$ (12.6\%) and $\text{F1}_{\Delta}^{\text{Age}}$ (7.1\%), indicating the poorest fairness in both gender and age dimensions. 
In contrast, GLoRIA obtained the highest fairness with an $\text{F1}_{\Delta}^{\text{Gender}}$ of {2.5\%} and $\text{F1}_{\Delta}^{\text{Age}}$ of {0.4\%}, respectively. 

\textbf{Effect of Fine-tuning on Fairness}. Fig.~\ref{fig:fairness}(d) presents the fairness metrics before fine-tuning. Comparison between Fig.~\ref{fig:fairness}(c)  and (d) shows that GLoRIA generally has reduced metric values (improved fairness) after fine-tuning. However, for the other models, the fine-tuning leads to higher unfairness. For example, for GLoRIA, $\text{EqOdds}^{\text{Gender}}$ decreased from 2.9\% to 1.4\%. For MedCLIP$^{\text{ViT}}$, $\text{EqOdds}^{\text{Gender}}$ and $\text{EqOdds}^{\text{Age}}$ increased from 2.3\% to 6.8\%
and from 3.0\% to 3.5\%, respectively. The comparison shows that the reduced fairness of the other models is unlikely due to our dataset, but from the models themselves.

\section{Conclusion}
In this work, we constructed a balanced chest X-ray image dataset to investigate the fairness of CLIP-like foundation models on different disease types and attributes including age and gender. Both utility and fairness of zero-shot inference and different strategies of fine-tuning are analyzed. Our main finding includes: 1) The zero-shot inference performance of these models is still low for chest X-ray image classification, and fine-tuning could significantly improve the performance. 2)  While MedCLIP$^{ViT}$ after full fine-tuning achieved the best performance, its fairness on different disease types and demographic attributes (age and gender) is lower than the other compared models, and GLoRIA has the best fairness metric values.  3) The effect of fine-tuning with a balanced dataset on fairness mainly depends on the model itself, with GLoRIA having improved fairness and the other models having decreased fairness after full fine-tuning, respectively. Our evaluation reveals a comprehensive understanding of performance and fairness in adapted CLIP-based models for chest X-ray image classification. 
These results underscore the necessity for developing advanced fairness interventions for foundation models to ensure equitable clinical diagnosis. 

\section{Compliance with Ethical Standards}
Ethical approval was not required as confirmed by the license attached with the open access data.

\section{Acknowledgment}
This work was supported by the National Key Research \& Development Program of China (2022ZD0160705).

\bibliographystyle{IEEEbib}
\bibliography{strings,refs}

\begin{thebibliography}{10}

\bibitem{ccalli2021deep}
Erdi Çallı, Ecem Sogancioglu, Bram {van Ginneken}, Kicky~G. {van Leeuwen}, and Keelin Murphy,
\newblock ``Deep learning for chest {X-ray} analysis: A survey,''
\newblock {\em Medical Image Analysis}, vol. 72, pp. 102125, 2021.

\bibitem{esteva2017dermatologist}
Andre Esteva, Brett Kuprel, Roberto~A Novoa, Justin Ko, Susan~M Swetter, Helen~M Blau, and Sebastian Thrun,
\newblock ``Dermatologist-level classification of skin cancer with deep neural networks,''
\newblock {\em Nature}, vol. 542, no. 7639, pp. 115--118, 2017.

\bibitem{qin2018computer}
Chunli Qin, Demin Yao, Yonghong Shi, and Zhijian Song,
\newblock ``Computer-aided detection in chest radiography based on artificial intelligence: a survey,''
\newblock {\em Biomedical engineering online}, vol. 17, pp. 1--23, 2018.

\bibitem{10635182}
Ugur Demir, Debesh Jha, and {Zheyuan Zhang,} et~al.,
\newblock ``Explainable transformer prototypes for medical diagnoses,''
\newblock in {\em ISBI}, 2024, pp. 1--5.

\bibitem{9761426}
Hoang Nguyen~Ngoc, Vu~Hoang, Trung~H. Bui, Steven Q.~H. Truong, Thanh~Huynh Minh, Duong Nguyen~Van, Trang Nguyen Thi~Minh, and Cong Cung~Van,
\newblock ``An efficient approach for tuberculosis diagnosis on chest {X-Ray},''
\newblock in {\em ISBI}, 2022, pp. 1--5.

\bibitem{10635205}
Dimitrios Pantelaios, Paraskevi-Antonia Theofilou, Paraskevi Tzouveli, and Stefanos Kollias,
\newblock ``Hybrid {CNN-ViT} models for medical image classification,''
\newblock in {\em ISBI}, 2024, pp. 1--4.

\bibitem{radford2021learning}
Alec Radford, Jong~Wook Kim, and {Chris Hallacy,} et~al.,
\newblock ``Learning transferable visual models from natural language supervision,''
\newblock in {\em ICML}, 2021, pp. 8748--8763.

\bibitem{8c537381a4af42e99177c9f1a94dc4cd}
Zifeng Wang, Zhenbang Wu, Dinesh Agarwal, and Jimeng Sun,
\newblock ``Medclip: Contrastive learning from unpaired medical images and text,''
\newblock in {\em EMNLP}, 2022, pp. 3876--3887.

\bibitem{huang2021gloria}
Shih-Cheng Huang, Liyue Shen, Matthew~P Lungren, and Serena Yeung,
\newblock ``Gloria: A multimodal global-local representation learning framework for label-efficient medical image recognition,''
\newblock in {\em ICCV}, 2021, pp. 3942--3951.

\bibitem{zhang2023biomedclip}
Sheng Zhang, Yanbo Xu, and {Naoto Usuyama,} et~al.,
\newblock ``Biomedclip: a multimodal biomedical foundation model pretrained from fifteen million scientific image-text pairs,''
\newblock {\em arXiv preprint arXiv:2303.00915}, 2023.

\bibitem{10230368}
Tochi Oguguo, Ghada Zamzmi, Sivaramakrishnan Rajaraman, Feng Yang, Zhiyun Xue, and Sameer Antani,
\newblock ``A comparative study of fairness in medical machine learning,''
\newblock in {\em ISBI}, 2023, pp. 1--5.

\bibitem{jin2024fairmedfm}
Ruinan Jin, Zikang Xu, Yuan Zhong, Qiongsong Yao, Qi~Dou, S~Kevin Zhou, and Xiaoxiao Li,
\newblock ``Fairmedfm: fairness benchmarking for medical imaging foundation models,''
\newblock in {\em NeurIPS Datasets and Benchmarks Track}, 2024, pp. 1--14.

\bibitem{wang2017chestx}
Xiaosong Wang, Yifan Peng, and {Le Lu,} et~al.,
\newblock ``Chestx-ray8: Hospital-scale chest {X-ray} database and benchmarks on weakly-supervised classification and localization of common thorax diseases,''
\newblock in {\em CVPR}, 2017, pp. 2097--2106.

\bibitem{irvin2019chexpert}
Jeremy Irvin, Pranav Rajpurkar, and {Michael Ko,} et~al.,
\newblock ``Chexpert: A large chest radiograph dataset with uncertainty labels and expert comparison,''
\newblock in {\em AAAI}, 2019, vol.~33, pp. 590--597.

\bibitem{alsentzer-etal-2019-publicly}
Emily Alsentzer, John Murphy, and {William Boag,} et~al.,
\newblock ``Publicly available clinical {BERT} embeddings,''
\newblock in {\em Clinical NLP}, 2019, pp. 72--78.

\bibitem{johnson2016mimic}
Alistair~EW Johnson, Tom~J Pollard, and {Lu Shen,} et~al.,
\newblock ``Mimic-iii, a freely accessible critical care database,''
\newblock {\em Scientific Data}, vol. 3, no. 1, pp. 1--9, 2016.

\bibitem{johnson2019mimic}
Alistair~EW Johnson, Tom~J Pollard, and {Seth J Berkowitz,} et~al.,
\newblock ``{MIMIC-CXR}, a de-identified publicly available database of chest radiographs with free-text reports,''
\newblock {\em Scientific Data}, vol. 6, no. 1, pp. 317, 2019.

\bibitem{gu2021domain}
Yu~Gu, Robert Tinn, and {Hao Cheng,} et~al.,
\newblock ``Domain-specific language model pretraining for biomedical natural language processing,''
\newblock {\em ACM Transactions on Computing for Healthcare (HEALTH)}, vol. 3, no. 1, pp. 1--23, 2021.

\bibitem{you2023cxr}
Kihyun You, Jawook Gu, Jiyeon Ham, Beomhee Park, Jiho Kim, Eun~K Hong, Woonhyuk Baek, and Byungseok Roh,
\newblock ``{CXR-Clip}: Toward large scale chest {X-ray} language-image pre-training,''
\newblock in {\em MICCAI}, 2023, pp. 101--111.

\bibitem{hu2022lora}
Edward~J Hu, Yelong Shen, and {Phillip Wallis,} et~al.,
\newblock ``Lora: Low-rank adaptation of large language models,''
\newblock in {\em ICLR}, 2022, pp. 1--13.

\end{thebibliography}

\end{document}